%% file: sample-sigconf.tex
\documentclass[sigconf]{acmart}
 
\usepackage{booktabs} 
\usepackage{amssymb,amsmath,subcaption}
\usepackage{algorithm}
\usepackage{algorithmic}
\usepackage{epstopdf}
\include{commands}

\usepackage{hyperref}
\usepackage{array}
\newcolumntype{L}{>{\centering\arraybackslash}m{3cm}}
\usepackage{wrapfig}
\usepackage{paralist}

\usepackage{graphics}    
\usepackage{graphicx}
\usepackage{color}

\setcopyright{rightsretained}

\begin{document}
\copyrightyear{2017} 
\acmYear{2017} 
\setcopyright{acmcopyright}
\acmConference{KDD '17}{}{August 13-17, 2017, Halifax, NS, Canada}
\acmPrice{15.00}\acmDOI{10.1145/3097983.3098038}
\acmISBN{978-1-4503-4887-4/17/08}

\fancyhead{}
\settopmatter{printacmref=false}

\title{Accelerating Innovation Through Analogy Mining}

\author{Tom Hope}
\affiliation{%
  \institution{The Hebrew University of Jerusalem}
}
\email{tom.hope@mail.huji.ac.il}

\author{Joel Chan}
\affiliation{%
  \institution{Carnegie Mellon University}
}
\email{joelchuc@cs.cmu.edu}

\author{Aniket Kittur}
\affiliation{%
  \institution{Carnegie Mellon University}
}
\email{nkittur@cs.cmu.edu}

\author{Dafna Shahaf}
\affiliation{%
  \institution{The Hebrew University of Jerusalem}
}
\email{dshahaf@cs.huji.ac.il}


\begin{abstract}
The availability of large idea repositories (e.g., the U.S. patent database) could significantly accelerate innovation and discovery by providing people with inspiration from solutions to analogous problems. However, finding useful analogies in these large, messy, real-world repositories remains a persistent challenge for either human or automated methods. Previous approaches include costly hand-created databases that have high relational structure (e.g., predicate calculus representations) but are very sparse. Simpler machine-learning/information-retrieval similarity metrics can scale to large, natural-language datasets, but struggle to account for structural similarity, which is central to analogy. 
In this paper we 
explore the viability and value of learning simpler structural representations, specifically, ``problem schemas'', which specify the purpose of a product and the mechanisms by which it achieves that purpose. Our approach combines crowdsourcing and recurrent neural networks to extract purpose and mechanism  vector representations from product descriptions. We demonstrate that these learned vectors allow us to find analogies with higher precision and recall than traditional information-retrieval methods. In an ideation experiment, analogies retrieved by our models significantly increased people's likelihood of generating creative ideas compared to analogies retrieved by traditional methods. Our results suggest a promising approach to enabling computational analogy at scale is to learn and leverage weaker structural representations.
%
%
\end{abstract}

%
%


\maketitle
 
\input{introduction}

\section{Learning a Representation for Analogies}
\input{Purp_mech_motivation}

\input{Purp_mech_formulation}

\section{Evaluation: Analogies}
\input{Analogy_distances_eval}

\section{Evaluation: Ideation by analogy}

\input{Analogy_ideation_eval}

\section{Discussion and Conclusion}
\input{Discussion}

\xhdr{Acknowledgments} The authors thank the anonymous reviewers for their helpful comments. This work was supported by NSF grants CHS-1526665, IIS-1149797, IIS-1217559, Carnegie Mellon's Web2020 initiative, Bosch, Google, ISF grant 1764/15 and Alon grant. Dafna Shahaf is a Harry\&Abe Sherman assistant professor.

\bibliographystyle{ACM-Reference-Format}
\bibliography{sigproc} 

\end{document}

%% file: commands.tex
\newcommand{\remove}[1]{}

\usepackage{amssymb}
\usepackage{graphicx}
\usepackage{color}
\newcommand{\xhdr}[1]{\vspace{1mm}\noindent{{\bf #1.}}}

%% file: introduction.tex
\section{Introduction}







The ability to find useful analogies is critical to driving innovation in a variety of domains.  Many important discoveries in science were driven by analogies: 
for example, an analogy between bacteria and slot machines helped Salvador Luria advance the theory of bacterial mutation. Analogical reasoning forms the foundation of law, with the effectiveness of an argument often dependent on its legal precedents   \cite{spellman2012legal}. Innovation is often spurred by analogy as well: an analogy to a bicycle allowed the Wright brothers to design a steerable aircraft. Whether architecture, design, technology, art, or mathematics, the ability to find and apply patterns from other domains is fundamental to human achievement \cite{hesse1966models, pask2003mathematics, markman2010structural, dahl2002influence}.


The explosion of available online data represents an unprecedented opportunity to find new analogies and accelerate human progress across domains. For example, the US Patent database has full text for more than 9 million patents issued from 1976 to the present. InnoCentive\footnote{innocentive.com} contains more than 40,000 business, social, policy, scientific, and technical problems and solutions. Quirky\footnote{quirky.com}, a company that assists inventors in the development process, has had over 2 million product idea submissions. OpenIDEO\footnote{OpenIDEO.com} receives hundreds of solutions for a variety of social problems. Millions of scientific papers and legal cases are searchable on Google Scholar. 

We believe these data form a treasure trove of analogies that can accelerate problem solving, innovation and discovery. In a striking recent example, a car mechanic invented a simple device to ease difficult childbirths by drawing an analogy to extracting a cork from a wine bottle, which he discovered in a YouTube video. This award-winning device could save millions of lives, particularly in developing countries. We imagine a future in which people could search through data based on deep analogical similarity rather than simple keywords; lawyers or legal scholars could find legal precedents sharing similar systems of relations to a contemporary case; and product or service designers could mine  myriad potential solutions to their problem. 

However, sifting through these massive data sources to find relevant and useful analogies poses a serious challenge for both humans and machines. In humans, memory retrieval is highly sensitive to surface similarity, favoring near, within-domain analogs that share object attributes over far, structurally similar analogs that share object relations \cite{gentner1985analogical,holyoak1996mental,gentner1993roles,gick_schema_1983}. Analogical processing also incurs a heavy cognitive load, taxing working memory when even a few relations are required to be processed at once \cite{halford2005many}. Thus searching through datasets with thousands or millions of items for structurally similar ones may be a daunting prospect.

Finding analogies is challenging for machines as well, as it is based on having an understanding of the deep relational similarity between two entities that may be very different in terms of surface attributes \cite{gentner1983structure}. For example, Chrysippus' analogy between sound waves and water waves required ignoring many different surface features between the two \cite{holyoak1996mental}. Recent advances in data mining and information retrieval include a variety of natural language techniques that use words, parts of speech or other language feature-based vector representations in order to calculate similarity measures (see \cite{shutova2010models}). Examples include word embedding models like Word2Vec \cite{mikolov_efficient_2013}, vector-space models like Latent Semantic Indexing \cite{deerwester_indexing_1990}, and probabilistic topic modeling approaches like Latent Dirichlet Allocation \cite{blei_latent_2003}. These approaches excel at detecting {\it surface similarity}, but are often unable to detect similarity between documents whose word distributions are disparate. The problem is especially acute when the source and target domains are different (for example, bacterial mutation and slot machines). 

Another approach to finding analogies has been to use the structural similarity of sentences or texts, such as using coupled clustering for detecting structural correspondence of text \cite{bollegala2009measuring, turney2006similarity}. However, these approaches typically require rich data sets with clear substructures, whereas most descriptions of problems or ideas in existing online databases are short, sparse, or lack consistent structure. Other current methods focus on very narrow analogy tasks, 
such as four-term analogy problems (teacher:student = doctor:?), in particular with short strings (ABC:ABD = KJI:?) \cite{hofstadter1994copycat}. In contrast, we wish to find analogies in real world data, which involve complex representations and a diverse set of analogical relations.


In this paper, we are interested in {\bf automatically discovering analogies in large, unstructured data sets}. In particular, we focus on a corpus of {\bf product innovations}.
There are two insights behind our approach that we believe may make this problem tractable despite its longstanding status as a ``holy grail'' in both cognitive science and AI.  First, rather than trying to solve the problem of fully structured analogical reasoning, we instead explore the idea that for retrieving practically useful analogies, we can use weaker structural representations that can be learned and reasoned with at scale (in other words, there is a tradeoff between the ease of extraction of a structure and its expressivity). Specifically, we investigate the weaker structural representation of an idea's {\it purpose} and {\it mechanism} as a way to find useful analogies. 
 The second insight is that advances in crowdsourcing have made it possible to harvest rich signals of analogical structure that can help machine learning models learn in ways that would not be possible with existing datasets alone. 

This paper combines these two ideas to contribute a technique for computationally finding analogies from unstructured text datasets that go beyond surface features. At a high level, our approach uses the behavioral traces of crowd workers searching for analogies and identifying the purpose and mechanisms of ideas, then developing machine learning models that develop similarity metrics suited for analogy mining. We demonstrate that learning purpose and mechanism representations allows us to find analogies with higher precision and recall than traditional information-retrieval methods based on TF-IDF, LSA, LDA and GloVe, in challenging noisy settings. Furthermore, we use our similarity metrics to automatically find \emph{far} analogies -- products with high purpose similarity, and low mechanism similarity. In a user study, we show that  we are able to ``inspire'' participants to generate more innovative ideas than alternative baselines, increasing the relative proportion of positively-rated ideas by at least $25\%$.

%% file: Purp_mech_motivation.tex
\subsection{Motivation}

Much work in computation analogy has focused on fully structured data, often with logic-based representations. For example \cite{FALKENHAINER19891}, 

\begin{verbatim}
CAUSE(GREATER-THAN[TEMPERATURE(coffee), 
      TEMPERATURE (ice-cube)], 
      FLOW(coffee, ice-cube, heat, bar)) 
\end{verbatim}

These representations, while very expressive, are notoriously difficult to obtain.
In this section, we investigate a weaker structural representation. Our goal is to come up with a representation that can be \emph{learned}, while still being expressive enough to allow analogical mining.

Analogies between product ideas are intricately related to their \textit{purpose} and \textit{mechanism}. Informally, we think of a product's purpose as ``what it does, what it is used for", and a product's mechanism is ``how it does it, how it works". The importance of a product's purpose and mechanism as core components of analogy are theoretically rooted in early cognitive psychology work on schema induction which define the core components of a schema as a goal and proposed solution to it (e.g., \cite{gick_schema_1983}). More recently, the practical value of defining a problem schema as a purpose and mechanism has been demonstrated to have empirical benefits in finding and using analogies to augment idea generation (e.g., \cite{yu2016distributed, yu2016encouraging, yu2014searching, yu2014distributed}).

Separating an idea into purpose and mechanisms enables core analogical innovation processes such as re-purposing: For a given product (such as a kitchen-sink cleaner) and its purpose, finding another way to put it to use (cleaning windows). To that end, assume (for the moment) that we have for each product $i$ two vectors, $\mathbf{p}_i$ and $\mathbf{m}_i$, representing the product's purpose and mechanism, respectively. Using this representation, we are able to apply rich queries to our corpus of products, such as:
{ \setdefaultleftmargin{2em}{3em}{}{}{}{}
\begin{compactitem}
\item \textit{Same purpose, different mechanism}. Given the corpus of all products $\mathcal{P}$, a product $i$ with (normalized) purpose and mechanism vectors $\mathbf{p}_i, \mathbf{m}_i$, and distance metrics $d_p(\cdot,\cdot), d_m(\cdot,\cdot)$ between purpose and mechanism vectors (respectively), solve:
\begin{equation}
\label{eq:same-purp}
\begin{aligned}
\underset{\mathbf{\tilde{i}}\in \mathcal{P} }{\text{argmin }} &   
d_p(\mathbf{p}_i,\mathbf{p}_{\tilde{i}}) \\
s.t.
&  d_m(\mathbf{m}_i,\mathbf{m}_{\tilde{i}}) \geq \text{threshold},
\end{aligned}
\end{equation}
\item \textit{Same Mechanism, different purpose}. Solve:
\begin{equation}
\label{eq:same-mech}
\begin{aligned}
\underset{\mathbf{\tilde{i}}\in \mathcal{P} }{\text{argmin }} &   
d_m(\mathbf{m}_i,\mathbf{m}_{\tilde{i}}) \\
s.t.
&  d_p(\mathbf{p}_i,\mathbf{p}_{\tilde{i}}) \geq \text{threshold},
\end{aligned}
\end{equation}
\end{compactitem}
}
The decomposition of products into purpose and mechanism 
also draws inspiration from engineering {\it functional models} and ontologies for describing products \cite{hirtz2002functional}. 
Although there is no set common definition of functions \cite{Ookubo07towardsinteroperability}, much research on functionality has been conducted in areas such as functional representation,
engineering design and value engineering. The scope of these ontologies, however, is highly ``mechanistic'' or engineering-oriented, while in many cases we observe in product data the purpose of a product is more naturally understood in others term  -- such as whether it is for entertainment, leisure, or more serious purposes, who is the target user (adults, children), and so forth. 

Importantly, our dataset of product descriptions contains noisy texts, often written informally by non-professional people. In these texts product descriptions are often lacking detail or are ill-defined. To automatically describe a product in terms of a formal functional model  would require an inordinate amount of meticulous data annotation and collection by professional engineers over a large number of product descriptions. 
We thus resort to a softer approach, hoping that a compromise on the level of detail will enable data-driven methods to automatically extract useful representations of product purpose and mechanism. 


Finally, we also make note of the potentially wider applicability of automatically extracting these representations from real-word product descriptions. Identifying the key components and functions of products could conceivably improve (or augment) search capabilities in internal or external product databases, and perhaps enhance recommender systems by better understanding what a user is looking for in a product and what a product offers. This last idea is connected to a line of work on ``product dimensions'' \cite{mcauley2013hidden}, in which it is shown that \textit{implicitly} identifying the properties of products (such as that Harry Potter is a book about wizards), helps in improving recommendations. The authors propose a method that combines ratings data with textual product reviews, hoping to implicitly recover topics in the text that inform recommendations. We too look at product dimensions, but target only two that are more abstract and broad, and directly learn them in a supervised fashion from annotated data.

\subsection{Data}

\xhdr{Innovation Corpus}
We test our approach with a corpus of product descriptions from Quirky.com, an online crowdsourced product innovation website.
Quirky is representative of the kinds of datasets we are interested in, because it is large (at the time of writing, it hosts upwards of 10,000 product ideas, of which our corpus included 8500), unstructured (ideas are described in natural language), and covers a variety of domains (invention categories) which makes cross-domain analogies possible. The following example illustrates the typical length and ``messiness'' of product ideas in this dataset:

{\footnotesize 
\begin{verbatim}
  Thirsty too
    Pet water bowl/dispenser for your vehicle cup holder.
    Over spill lip to catch water
    Has optional sleeve for larger cup holders
    Optional floor base
    One way valve so water cant over flow from bottle
    Small reservoir 
    Reservoir acts as backsplash
    Water bottle attachment
    Holds water in your vehicle cupholder for pet
    Foldable handle to get unit out of holder
    Dishwasher safe
    Optional sleeve for larger cup holders
\end{verbatim}}

\begin{figure}
    \setlength{\belowcaptionskip}{-10pt}
	\includegraphics[width=1.1\linewidth]{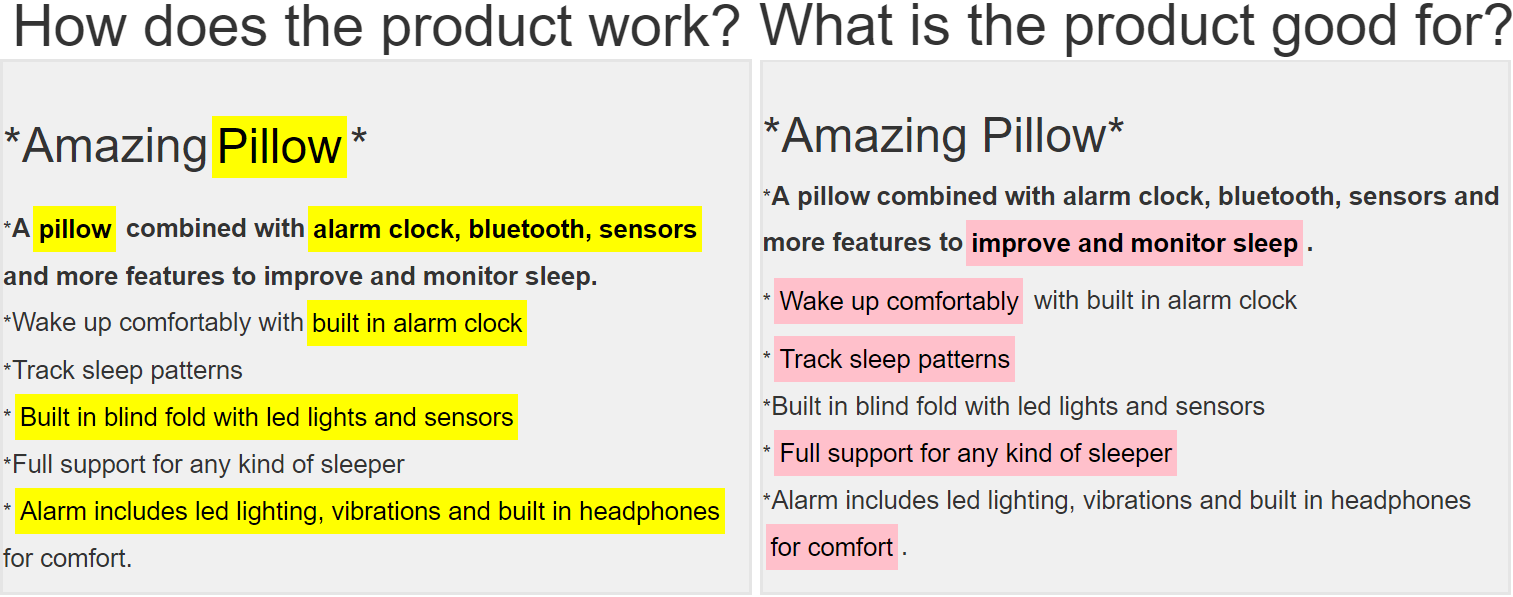}
	\caption{\small Collecting purpose and mechanism annotations from the crowd.\label{fig:purposemechcollect}}
\end{figure}

\xhdr{Collecting Purpose and Mechanism Data}
In addition to the Quirky innovation corpus, we needed to collect analogy-specific data to train our model. Previous approaches to creating structured representations of items for analogical computation, for example predicate logic, are extremely heavyweight and can take tens of person-hours for complex items \cite{vattam_dane:_2011}. Instead, we aim to develop a lightweight task that avoids complex structure but instead relies on the cognitive expertise and intuitions of people to be able to separate the purpose of a product from its mechanism.  By doing so we can scale up the collection of purpose and mechanism labels through the use of microtasks and crowdsourcing markets \cite{kittur_crowdsourcing_2008}.
%
Specifically, we show Amazon Mechanical Turk (AMT) crowd workers a product description, asking them to annotate the parts of the text they consider to be about the purposes of the product, and the parts related to mechanisms. We frame the problem in simple terms, guiding workers to look for words/phrases/chunks of text talking about ``what the product does, what it is good for" (purposes), and ``how it works, what are its components" (mechanisms). As seen in Figure \ref{fig:purposemechcollect}, we juxtapose two copies of the product text side-by-side, to ease cognitive load and encourage workers not to give purpose and mechanism tags that are too similar or overlapping, thus capturing a potentially richer and more distinct signal. Our corpus consisted of $8500$ products. Each product was annotated by four workers.
 
\xhdr{Collecting Analogies}
In previous, preliminary work \cite{chan2016scaling}, we explored the use of crowdsourcing to label analogies, collecting labeled examples of analogies (product pairs) that were fed into a metric-learning deep learning model. While showing promising results, the process of collecting labeled analogies proved expensive, requiring considerable cognitive effort from workers and thus more time, limiting the number of labels that can realistically be collected. In addition, in that work, the deep learning model was blind to the rich structures of purpose and mechanism, and had no hope of recovering them automatically due to relative data scarcity. In this paper  we take a different approach, focusing our resources on collecting purpose and mechanism annotations from the crowd, while collecting only a small number of curated labeled analogies strictly for the purpose of evaluation (see Section \ref{sec:evalpm} for details).

%% file: Purp_mech_formulation.tex
\subsection{Method}
\subsubsection{Extracting Purpose and Mechanism vectors}
In this section, we describe our approach to learning to extract purpose and mechanism product representations. We begin with a set of $N$ training product texts $\mathcal{X}_N = \{ \mathbf{x}_1, \mathbf{x}_2,\ldots, \mathbf{x}_N \}$, where each $\mathbf{x}_i$ is a variable-length sequence of tokens $({x}^1_i,{x}^2_i,\ldots,{x}^T_i)$. For each document $\mathbf{x}_i$, we collect a set of $K$  purpose annotations and $K$ mechanism annotations, where $K$ is the number of workers who annotate each document. We define the \textit{purpose annotation} to be a binary vector

$$\mathbf{\tilde{p}}_{i_k} = ({\tilde{p}}^1_{i_k},{\tilde{p}}^2_{i_k},\ldots,{\tilde{p}}^T_{i_k})$$ 

of the same length as $\mathbf{x}_i$, with $\tilde{p}^j_{i_k} = 1$ if token ${x}^j_i$ is annotated as purpose by annotator $k$, $\tilde{p}^j_{i_k} = 0$ if not. In the same way, we denote the \textit{mechanism annotation} with  
$$\mathbf{\tilde{m}}_{i_k} = ({\tilde{m}}^1_{i_k},{\tilde{m}}^2_{i_k},\ldots,{\tilde{m}}^T_{i_k})$$ 

While on the surface this setting appears to lend itself naturally to sequence-to-sequence learning \cite{sutskever2014sequence}, there are a few important differences. A key difference is that in our setting the problem of interest is not to learn to recover the latent  $\mathbf{\tilde{p}}$ ($\mathbf{\tilde{m}}$) exactly for unseen products, but rather to extract some form of representation that captures the \textit{overall} purpose and mechanism. 
That is, what we do care about is the semantic meaning or context our representation captures with respect to product purposes or mechanisms, rather than predicting any individual words. Additionally, sequence-to-sequence models typically involve heavier machinery and work well on large data sets, not suitable for our scenario were we have at most a few thousands of tagged examples. 

On a more technical note, instead of one sequence in the output, we now have $K$. A simple solution is to aggregate the annotations, for example by taking the union or intersection of annotations, or considering a token ${x}^j_i$ to be positively annotated if it has at least $\frac{K}{2}$ positive labels. Richer aggregations may also be used.

Considering that our focus here is to capture an overall representation (not predict a precise sequence), however, we resort to a simple and soft aggregation of the $K$ annotations. In simple terms, we look at all words that were annotated, and take a TF-IDF-weighted average of their word vectors. 

In more formal terms, let $\mathbf{w}_i = ({w}^1_i,{w}^2_i,\ldots,{w}^T_i)$ be the sequence of GloVe \cite{pennington2014glove} word vectors (pre-trained on Common Crawl web data), representing $({x}^1_i,{x}^2_i,\ldots,{x}^T_i)$. We select all ${x}_i$ word vectors for which $\tilde{p}^j_{i_k}=1$ ($\tilde{m}^j_{i_k}=1$) for some $k$, and concatenate them into one sequence. We then compute the TF-IDF scores for tokens in this sequence, find the $D$ tokens with top TF-IDF scores ($D=5$ in our experiments), and take the TF-IDF-weighted average of their corresponding $D$ GloVe vectors. We denote the resulting weighted-average vectors as $\mathbf{p}_i \in \mathbb{R}^{300}$ and $\mathbf{m}_i \in \mathbb{R}^{300}$ for purpose and mechanism annotations, respectively. We consider $\mathbf{p}_i, \mathbf{m}_i$ to be \textit{target} vectors we aim to predict for unseen texts.

Embedding (short) texts with a weighted-average of word vectors (such as with TF-IDF) can lead to surprisingly good results across many tasks \cite{arora2017embed}. Furthermore, in our case this simple weighted-average has several advantages. As we will next see, it lends itself to a straightforward machine learning setting, suitable for our modestly-sized data set, and for the objective of finding an overall vector representation that can be used in multiple ways, chiefly the computation of \textbf{purpose-wise and mechanism-wise distances} between products. Additionally, by concatenating all annotations and weighting by TF-IDF, we naturally give more weight in the average vector to words that are more frequently annotated -- thus giving higher impact to words considered important by all annotators with respect to purpose/mechanism.

\subsubsection{Learning purpose and mechanism}
We now have $N$ training product texts $\mathcal{X}_N = \{ \mathbf{x}_1, \mathbf{x}_2,\ldots, \mathbf{x}_N \}$, and $N$ corresponding target tuples $\mathcal{Y}_N = \{ (\mathbf{p}_1,\mathbf{m}_1), (\mathbf{p}_2,\mathbf{m}_2),\ldots, (\mathbf{p}_N,\mathbf{m}_N) \}$. We represent each $\mathbf{x}_i$ with its pre-trained GloVe vectors, $\mathbf{w}_i$. Our goal is to learn a function $f(\mathbf{w}_i)$ that  predicts $(\mathbf{p}_i,\mathbf{m}_i)$. To this end, we model $f(\cdot)$ with a Recurrent Neural Network as follows. The network takes as input the variable-length sequence $\mathbf{w}_i$. This sequence is processed with a bidirectional RNN (BiRNN) \cite{bahdanau2014neural} with one GRU layer. The BiRNN consists of the forward
GRU $\overrightarrow{GRU}$ which reads the sequence $\mathbf{w}_i$ from ${w}^1_i$ to ${w}^T_i$, and a backward GRU $\overleftarrow{GRU}$ which reads from ${w}^T_i$
to  ${w}^1_i$, thus in practice capturing the neighborhood of ${w}^j_i$ from ``both directions":
$$
\begin{aligned}
 \overrightarrow{{h}^j_i} = \overrightarrow{GRU}({w}^j_i), \  
 \overleftarrow{{h}^j_i}  = \overleftarrow{GRU}({w}^j_i), \ 
 {h}^j_i = [\overrightarrow{{h}^j_i}, \overleftarrow{{h}^j_i}],
\end{aligned}
$$
where we concatenate forward and backward GRU hidden states to obtain ${h}^j_i$, our representation for word $j$ in product $i$. In our case, we are interested in ${h}^T_i$ which captures the entire product text. 

Next, let $\mathbf{W}_p$ and $\mathbf{W}_m$ be \textbf{purpose and mechanism weight matrices}, respectively. ${h}^T_i$ is a \textit{shared} representation of the document, which we now transform into two new vectors, $\mathbf{\hat{p}}_i, \mathbf{\hat{m}}_i$, forming our purpose and mechanism predictions for product $i$:

\begin{equation}
\label{eq:proj}
\begin{aligned}
 \mathbf{\hat{p}}_i = \mathbf{W}_p {h}^T_i, \  
 \mathbf{\hat{m}}_i = \mathbf{W}_m {h}^T_i.  
\end{aligned}
\end{equation}

Parameters in this network are then tuned to minimize the $MSE$ loss averaged over $(\mathbf{p}_i,\mathbf{m}_i)$. In some scenarios, we may care more about predicting either purpose or mechanism, and in that case could incorporate a weight term in the loss function, giving more weight to either $\mathbf{p}_i$ or $\mathbf{m}_i$.

\subsubsection{Purpose and Mechanism vector interpretations} 
Here, we give intuition about the kinds of representations extracted and the ability to interpret them with very simple tools. We first compute $\mathbf{\hat{p}}, \mathbf{\hat{m}}$ for held-out product texts. Then, in the first approach to interpreting purpose and mechanism predictions, we find the top $10$ GloVe word vectors $w$ most similar to each of $\mathbf{\hat{p}}, \mathbf{\hat{m}}$, among all vectors that appear in our vocabulary.

\begin{table*}[t]
	\caption{\small Purpose and Mechanism vector interpretation examples. Descriptions shortened. Sparse coding shows only words with $|\alpha| \geq 0.1$ }
	\label{table:interp_vecs}
	{\small
    \setlength{\parindent}{0em}
\setdefaultleftmargin{1em}{2em}{}{}{}{}
		\begin{tabular}{|p{.25\linewidth}|p{.25\linewidth}|p{.25\linewidth}|}
			\hline	
			Product  & Purpose words & Mechanism words \\
			\hline
			A small  yogurt maker machine for concentrating yogurt under heat and vacuum.  Has a round base in drum with customized scooper,  washable stainless steel drum parts. Reduce time and energy used.   &  \textbf{Top similar:} concentrate, enough, food, even, much, especially, reduce, produce, whole
            
\textbf{Sparse coding:} making, energy, yogurt, drum, 
concentrate, vacuum, heavy, foods, aches, service
& 
\textbf{Top similar:} liquid, heat, cooling, pump,
steel, machine, water, heating, electric
            
\textbf{Sparse coding:} vacuum, cooled, drum,
             heavy, ingredients, design, renewable, stainless, vending		
\\ \hline
A cover placed on a car truck to protect from hail. Elastic perimeter to prevent wind from blowing under cover. Snap or velcro slits to open door without removing cover. Strong attachment so it won’t blow away. Inflatable baffles that cover the top, front windshield, side. 
& 
\textbf{Top similar:} storm, hail, rain, roofs,
doors, wind, front, winds, walls

\textbf{Sparse coding:} roof, hail, padded,
obstructing, defenses, diesel, windshield, wets
&
\textbf{Top similar:} roof, cover, lining, zipper,
bottom, hood, plastic, flap, rubber 

\textbf{Sparse coding:} front, cover, insulation,
hail, buckle, sling, watertight, cutter, blowing
\\ \hline
A leash accessory with removable compartments for phone, cards cash, keys, poop bags, treats, bowl.  Walk your dog and carry your essentials without pockets or a purse bag.
& 
 \textbf{Top similar:} bags, purse, wallet, carry, leash,
 backpack, pocket, dog, luggage

\textbf{Sparse coding:} bag, leash, compartments,
pets, phone, eats, practical, handing, pull        
& 
\textbf{Top similar:} leash, pouch, purse, pocket, 
bags, pockets, strap, compartment, backpack

\textbf{Sparse coding:} leash, bag, compartments, hand, holders
\\ \hline
					
\end{tabular}
	}
\end{table*}

In the second approach, we aim to recover a set of $10$ word vectors such that their sparse linear combination approximately gives  $\mathbf{\hat{p}}$ or $\mathbf{\hat{m}}$. More formally, in the spirit of the sparse coding approach in \cite{arora2016linear}, consider the collection of all word vectors in our vocabulary $V$, ${w}_1,{w}_2,\ldots,{w}_{|V|}$. We stack them into a matrix $\mathbf{W}$. We aim to solve the following optimization problem:

\begin{equation}
\label{eq:sparse}
\begin{aligned}
\underset{\mathbf{a}}{\text{argmin}} & ||\mathbf{\hat{p}}_i - \mathbf{W}\mathbf{a}||^2_2 
\ \ \ s.t. \ \ 
& ||\mathbf{a}||_0 \leq 10,
\end{aligned}
\end{equation}

where $\mathbf{a}$ is a weight vector. Optimization  can be done with the Orthogonal Matching Pursuit (OMP) \cite{cai2011orthogonal} greedy algorithm. 

In Table \ref{table:interp_vecs}, we display some examples of applying these two simple methods, to product texts in test data (not seen during training). The first product is a yogurt maker machine, \textit{used for} concentrating yogurt under heat, and to \textit{reduce} time and energy. We observe that words selected as most related to our purpose vector representation include \textit{food, produce, concentrate, making, energy, reduce} and also words that are typical in the language used in describing advantages of products in our data, such as \textit{especially, whole, enough, much}. Mechanism words, are indeed overall of a much more ``mechanical'' nature, including \textit{liquid, heat, cooling, pump, steel, machine}. In the other examples too, we observe the same pattern: Words selected as most  closely-related to purpose or mechanism representations, using simple techniques, empirically appear to reflect corresponding properties in the product text, both in language and in deeper meaning.

%% file: Analogy_distances_eval.tex
As typically done in the context of learning document representations, the key approach to quantitative evaluation is a down-stream task such as document classification. We now evaluate the predicted $\mathbf{\hat{p}}_i ,\mathbf{\hat{m}}_i$ in the context of their ability to capture distances that reflect analogies, which is the primary focus of this paper.
To do so, we first create a dataset of analogies and non-analogies.

\subsection{Collecting analogies via crowdsourcing}
\label{sec:evalpm}



\begin{figure*}[t!]
\centering
\setlength{\belowcaptionskip}{-10pt}
\includegraphics[width=1.7\columnwidth]{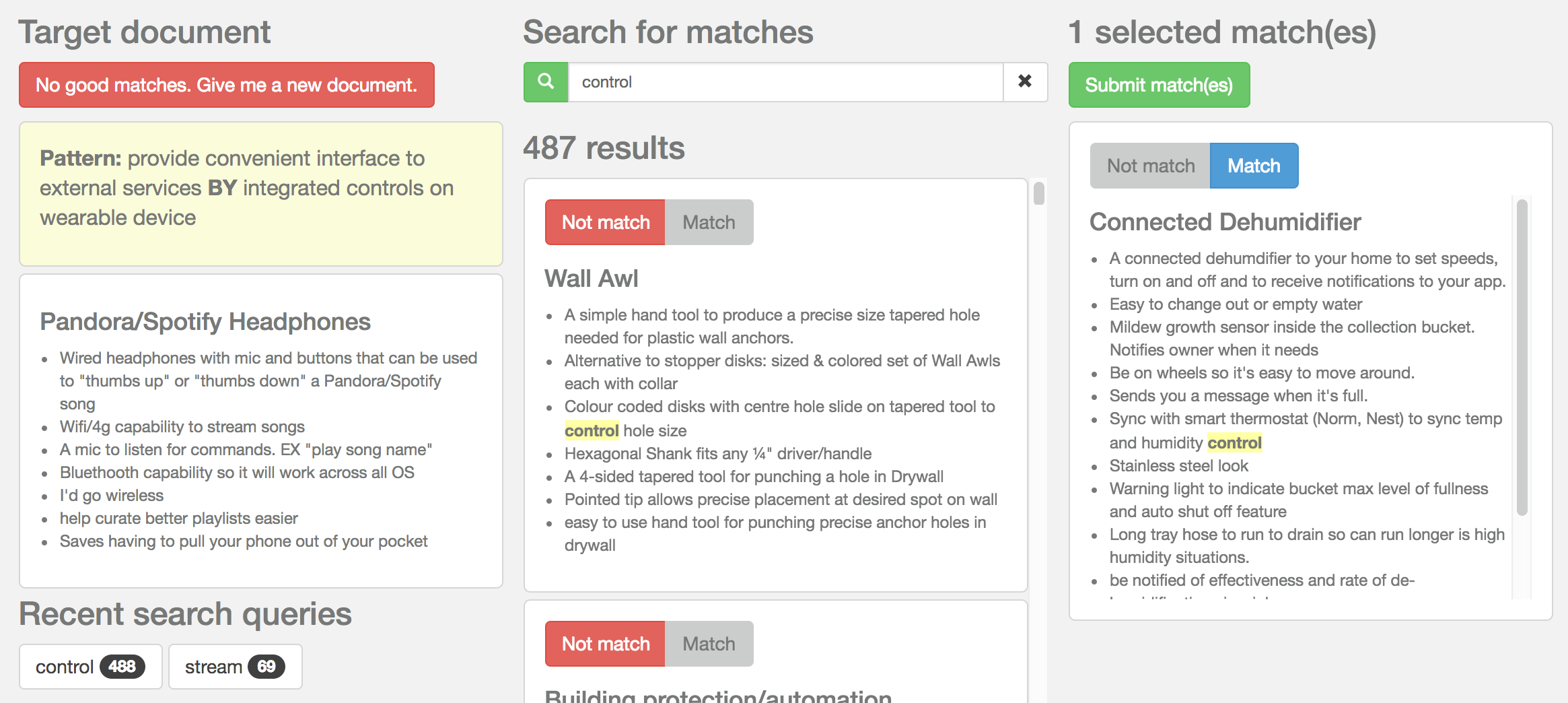}
\caption{Screenshot of analogy search interface}
\label{fig:search-ui}
\end{figure*}



We crowdsourced analogy finding within a set of about $8000$ Quirky products. AMT crowd workers used our search interface to collect analogies -- pairs of products -- for about $200$ seed documents. The search task is powered by a simple word-matching approach. To deal with word variants, we added lemmas for each word to the bag-of-words associated with each product. Each search query was also expanded with lemmas associated with each query term. Search results were ranked in descending order of number of matching terms. Median completion time for each seed was 7 minutes (workers could complete as many seeds as they wanted). Further, to deal with potential data quality issues, we recruited 3 workers for each seed (to allow for majority-vote aggregation). 

Pairs that were tagged as \textbf{matches} became positive examples in our analogy dataset. However, coming up with negative examples was more difficult. Borrowing from information retrieval, we assume that people read the search results sequentially, and treat the \textbf{implicitly rejected} documents (i.e., documents that were not matches, despite appearing before matches) as negatives. It is important to remember that these documents are not necessarily real negatives. To further increase the chance that the document has actually been read, we restrict ourselves to the top-5 results.


\xhdr{Challenges}
Getting workers to understand the concept of analogies and avoiding tagging products that are superficially similar (e.g., ``both smartphone-based'') as analogies proved a challenge. To address this, we scaffolded the search task by first requiring workers to generate a schema (or ``pattern'') to describe the core purpose and mechanism of the product, first in concrete terms, and then in more abstract terms (see an example pattern Figure \ref{fig:search-ui}).
Workers were then instructed to find other products that matched the abstract schema they created. We found that this scaffolded workflow reduced the number of superficial matches; yet, a non-negligible portion of the pairs labeled as positive were either superficial matches or near analogies (i.e., analogies with many shared surface features), likely due to the strong tendency towards surface features in analogical retrieval \cite{gentner1993roles}. Further, because products were multifaceted, search results may have been implicitly rejected even if they were analogous to the seed if the matching schema was different from the one initially identified by the worker. 





\subsection{Quantitative results}

In Table \ref{table:model_results}, we present precision and recall @ K results. We rank all pairs in the test data ($N=2500$, with training done on about $5500$ products) based on their distances, according to various metrics, including our own. In summary, across all levels our approach outperformed the baselines, despite a challenging noisy setting. A considerable portion of test product pairs were tagged by workers as analogies despite having only surface similarity, creating mislabeled positive examples that favor the surface-based baselines.
 In addition to ranking by purpose-only and mechanism-only, we also concatenate both representations in a vector $[\mathbf{p}_i,\mathbf{m}_i]$ for product $i$, and observe an overall improvement in results, although the ``one-dimensional'' use of either purpose or mechanism alone still beats the baselines. Using $\mathbf{m}_i$ only led to considerably better results when looking at precision @ top $1\%$, perhaps indicating a tendency by workers to find more mechanism-based analogies.

\begin{table*}[t]
	\centering
	\caption{\small \textbf{Model results.} Precision, recall of positive labels @ top-scoring pairs (ranked by similarity). }
	\label{table:model_results}
	{\small
		\begin{tabular}{|c|c|c|c|c|c|c|}
			\hline	
			Method &Top $1\% $  & Top $5\%$ & Top $10 \% $ & Top $15 \% $ & Top $20\%$  & Top $25\%$ \\
			\hline
			Glove + TF-IDF (top 5 words) & 0.565, 0.018 & 0.515, 0.081 & 0.489, 0.153& 0.468, 0.22 & 0.443, 0.277 & 0.434, 0.339 \\ \hline
			Glove + TF-IDF  & 0.609, 0.019  & 0.559, 0.087 &  0.487, 0.152  & 0.47, 0.22& 0.449, 0.281 &  0.426 ,0.332\\ \hline
			TF-IDF & 0.63,0.02 & 0.537, 0.084 & 0.5 0.156 & 0.468, 0.22 & 0.464, 0.29&  0.441, 0.344\\ \hline
			LSA & 0.413, 0.013 & 0.463, 0.072 & 0.446, 0.14 & 0.435, 0.204 & 0.413, 0.258 & 0.399, 0.312 \\ \hline
           LDA & 0.435, 0.014 & 0.432, 0.067 & 0.414, 0.129 & 0.398, 0.187 & 0.384, 0.24 & 0.381, 0.298 \\ \hline

			Purpose only & 0.674, 0.021  & 0.586, 0.092
  & 0.535, 0.167  & 0.505, 0.237 & 0.496, 0.31 & 0.465, 0.363  \\ \hline
            Mechanism only &\textbf{0.739}, 0.023  & 0.586, 0.092 &0.551, 0.172
  &\textbf{0.507}, 0.237 &0.482, 0.301 & 0.47, 0.368 \\ \hline
            concat(Purpose, Mechanism) &0.696, 0.022 & \textbf{0.612}, 0.096 & \textbf{0.555}, 0.173  & \textbf{0.507}, 0.237  & \textbf{0.504}, 0.315 & \textbf{0.478}, 0.373  \\ \hline
    
		\end{tabular}
	}
\end{table*}

%% file: Analogy_ideation_eval.tex
Since a major application of the enhanced search and retrieval capabilities of analogy is enhanced creativity, we now evaluate the \textbf{usefulness} of our algorithms. We examine the degree to which our model's retrieved output improves people's ability to generate creative ideas, compared to other methods. To do so we use a standard ideation task in which participants redesign an existing product \cite{ullman_mechanical_2002}, and are given inspirations to help them -- either from our approach, a TF-IDF baseline, or a random baseline. 

See Figure \ref{fig:ideation} for an example task given to crowdworkers. Here, the task was to redesign a cell phone case that can charge the phone. The middle part shows the top 3 inspirations per condition. 

Our assumption is that our approach will provide participants with useful examples that are similar in purpose but provide diverse mechanisms that will help them explore more diverse parts of the design space in generating their ideas. We hypothesize that this approach will lead to better results than the TF-IDF baseline (highly relevant but non-diverse inspirations, focusing on surface features) and the random baseline (highly diverse but low relevance).

\subsection{Generating near-purpose far-mechanism analogies}

To generate inspirations for the redesign task, we start by using the learned purpose and mechanism representations $\mathbf{p}_i,\mathbf{m}_i$ for each document $i$ (in the test set) to apply rich queries to our corpus of products. In particular, assuming all vectors are normalized to unit euclidean norm, we can find pairs of products $i_1,i_2$ such that  $d_p(\mathbf{p}_{i_1},\mathbf{p}_{i_2}) = \mathbf{p}_{i_1}\cdot\mathbf{p}_{i_2}$ is high (near purpose), while $d_m(\mathbf{m}_{i_1},\mathbf{m}_{i_2}) = \mathbf{m}_{i_1}\cdot\mathbf{m}_{i_2}$ is low (far mechanism). This type of reasoning, as discussed above, is a core element of analogical reasoning. 

We take this idea one step forward by \textit{clustering} by purpose and \textit{diversifying} by mechanism. In more detail, we take a set of $2500$ products not seen during training, and follow a simple and intuitive procedure as follows. Let $\mathcal{P}_T$ denote our corpus of test-set products. Let $S$ denote the number of seed products we wish to use in our experiment. Let $M$ denote the number of inspirations we wish to produce for each seed $\{1,\ldots,P\}$.

\xhdr{Clustering by purpose} First, we find groups of products with similar purpose by clustering by our purpose representation.
{ \setdefaultleftmargin{2em}{3em}{}{}{}{}
\begin{compactitem}
	\item  Run K-means ($K=50$), based on vectors $\mathbf{p}_i, \forall i \in\mathcal{P}_T$. (Note that when all vectors are normalized, the euclidean norm on which K-means is based is equivalent to the cosine distance).
	\item  For each cluster $k \in \{1,\ldots,K \}$, compute an intra-distance measure (purpose homogeneity) $d_k$. We use the $MSE$. Prune clusters with less than $M$ instances. Rank clusters by $d_k$ in descending order, pick top $P$. Call this set of clusters $\mathcal{K}_\text{top-purpose}$, with corresponding cluster centers $\bar{\mathbf{p}}_1,\ldots,\bar{\mathbf{p}}_S$.
	\item For each cluster $k$ in $\mathcal{K}_\text{top-purpose}$, select the product $i$ whose vector $\mathbf{p}_i$ is nearest to the cluster center $\bar{\mathbf{p}}_k$. This is our $k^{\text{th}}$ \textbf{seed product}, denoted by $s_k$.
\end{compactitem}
}
\xhdr{Result diversification by mechanism} We now have a set of seed products, each with a corresponding cluster of products with similar purposes. Next, we need to pick $M$ inspirations per seed. 

For each seed $s_k$, we have a set of \textit{candidate matches} $\mathcal{U}_{s_k}$, all from the same purpose cluster. We empirically observe that in the purpose-clusters $\mathcal{K}_\text{top-purpose}$ we generate, some vectors are highly similar to the seed  with respect to mechanism, and some less so. In order to generate far-mechanism results for each seed from candidate set $\mathcal{U}_{s_k}$, we now turn to diversification of results. 

The problem of extracting a well-diversified subset of results from a larger set of candidates has seen a lot of work, prominently in the context of information retrieval (which is closely related to our setting). In our case, we assume to have found a set of relevant results $\mathcal{U}_{s_k}$ according to purpose metric $d_p(\cdot,\cdot)$, and diversify by mechanism metric  $d_m(\cdot,\cdot)$. 

There are many ways to diversify results, mainly differing by objective function and constraints. Two  canonical measures are the MAX-MIN and MAX-AVG dispersion problems \cite{ravi1994heuristic}. In the former, we aim to find a subset $\mathcal{M}\subseteq \mathcal{U}_{s_k}$ such that $|\mathcal{M}|=M$, and
 $$\min_{\mathbf{m}_{i_1},\mathbf{m}_{i_2} \in \mathcal{M}} d_m(\mathbf{m}_{i_1},\mathbf{m}_{i_2})$$ 
is maximized. In the latter, we aim to find a subset $\mathcal{M}\subseteq \mathcal{U}_{s_k}$ such that
$|\mathcal{M}|=M$, and
$$\frac{2}{M(M-1)}\sum_{\mathbf{m}_{i_1},\mathbf{m}_{i_2} \in \mathcal{M}} d_m(\mathbf{m}_{i_1},\mathbf{m}_{i_2}) $$
is maximized.
In other words, in the MAX-MIN problem we find a subset of products $\mathcal{M}$ such that the distance between the two nearest products is maximized. In the MAX-AVG problem, we find a subset such the average distance between pairs is maximized. Both problems admit simple greedy algorithms with constant-factor approximations \cite{ravi1994heuristic}. We choose the MAX-MIN problem, since we want to avoid displaying too-similar results even once to a user (who may become frustrated and not proceed to read more inspirations). 

We solve the problem using the GMM algorithm mentioned in \cite{ravi1994heuristic}. Each iteration of GMM selects a candidate $m \in  \mathcal{U}_{s_k} - \mathcal{M}$ such that the minimum distance from $m$ to an already-selected product in  $\mathcal{M}$ is the largest among remaining candidates in $\mathcal{U}_{s_k} - \mathcal{M}$, where we measure distance according to our mechanism metric $d_m(\cdot,\cdot)$.

In our experiments, we set $P=12, M=12$, for $12$ seeds and $12$ matches each, respectively.

\subsection{Experiment design}
We recruited 38 AMT workers to redesign an existing product, a common creative task in design firms \cite{ullman_mechanical_2002}. To ensure robustness of effects, the experiment included 12 different ``seed'' products. 
Participants were paid \$1.5 for their participation.

To maximize statistical power, we utilized a within-subjects design with a single manipulated factor, \textit{inspiration\_type}: 
{ \setdefaultleftmargin{2em}{3em}{}{}{}{}
\begin{compactitem}
\item ANALOGY: participants receive $12$ product inspirations retrieved by our method detailed above, using near-purpose far-mechanism clustering and diversification.
\item BASELINE: SURFACE: participants receive product inspirations retrieved using TF-IDF, by finding the top $12$ products similar to the seed. This baseline is meant to simulate current search engines.  
\item BASELINE: RANDOM: participants receive $12$ product inspirations sampled at random from our product corpus. 
\end{compactitem}}

Since we used a within-subjects design, participants completed the redesign task under each of the 3 inspiration\_type conditions. The order of conditions was counterbalanced to prevent order effects. To ensure unbiased permutations, we used the Fisher-Yates shuffle to assign seeds to conditions, so that every seed would be seen in all conditions (by different users).

\begin{figure}[t]
\centering 
	\setlength{\belowcaptionskip}{-10pt}
 	\includegraphics[width=\linewidth]{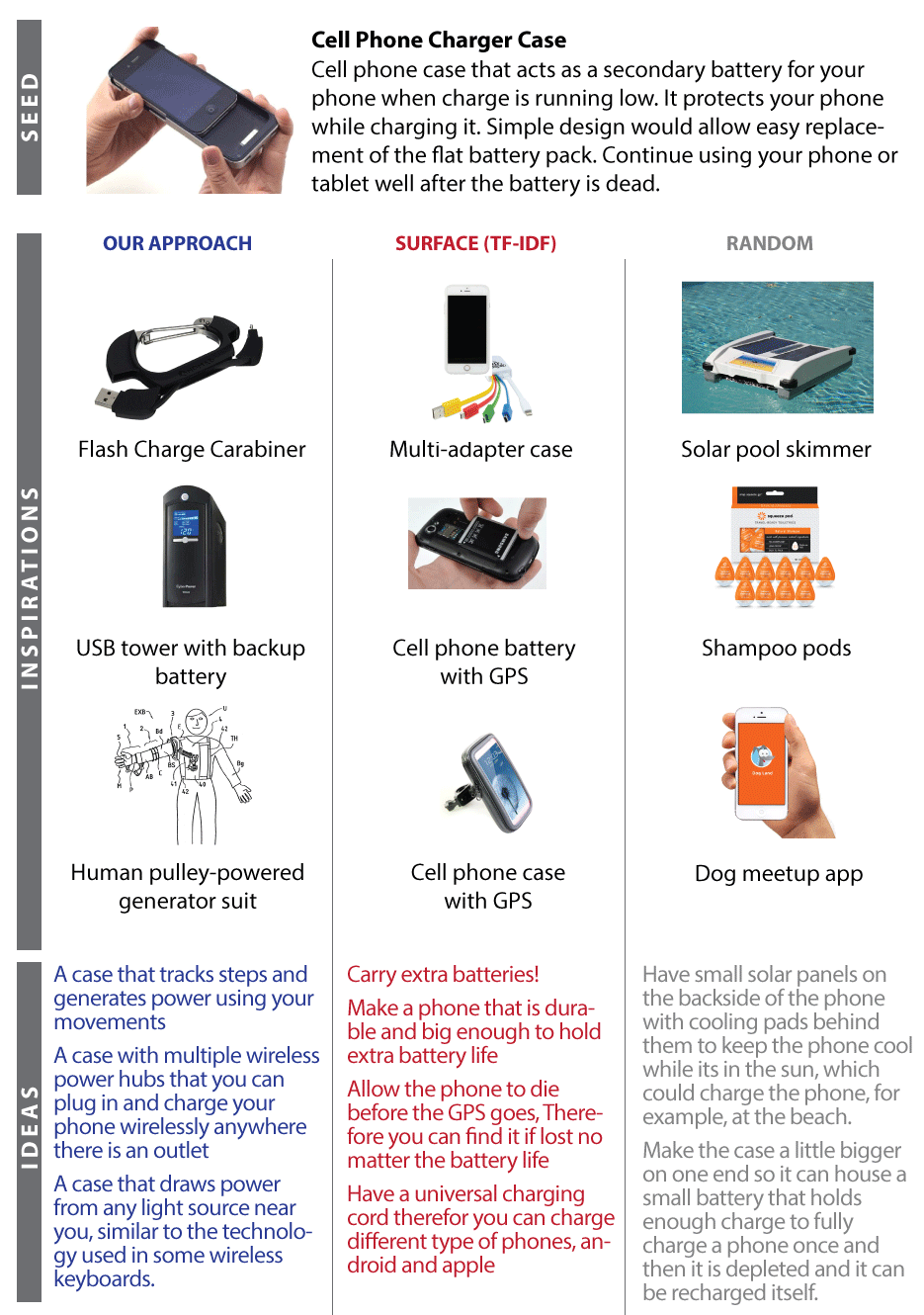} 
 	\caption{\small Overview and excerpts of the ideation experiment. Top: Seed product. Workers were asked to solve the same problem in a different way. Middle: Top 3 inspirations for each of the conditions. Note that the TF-IDF baseline returns results from the same domain, while our method returns a broader range of products. Bottom: Ideas generated by users exposed to the different conditions. \label{fig:ideation}}
\end{figure}

Since prior work has shown that people benefit more from analogies if they receive them after ideation has begun \cite{tseng_role_2008}, the ideation task proceeded in two phases: 1) generating ideas unassisted for one minute, then 2) receiving 12 inspirations and generating more ideas for 6 minutes. The inspirations were laid out in four pages, 3 inspirations per page, and the users could freely browse them. 


Figure \ref{fig:ideation} provides an overview of the experiment and an excerpt from the data. The original task was to redesign an existing product, in this case a cell phone charger case. The SURFACE baseline retrieves products that are very phone-related (or case related). In contrast, our algorithm retrieves diverse results such as a human pulley-powered electricity generator suit. The bottom of the figure shows ideas generated by users in each condition. Interestingly, the user exposed to our approach suggested a case that generates power using movement, potentially inspired by the suit.
 
\subsection{Results}
\xhdr{Measures}
We are interested in the ability of our approach to enhance people's ability to generate creative ideas. Following \cite{reinig_measurement_2007}, we measured creative output as the rate at which a participant generates good ideas. We recruited five graduate students to judge each idea generated by our participants as good or not. Our definition of ``good'' follows the standard definition of creativity in the literature as a combination of novelty, quality, and feasibility \cite{runco_standard_2012}. Each judge was instructed to judge an idea as good if it satisfied all of the following criteria: 1) it uses a different mechanism/technology than the original product (novelty), 2) it proposes a mechanism/technology that would achieve the same purpose as the original product (quality), and 3) could be implemented using existing technology and does not defy physics (feasibility).

Agreement between the judges was substantial, with a Fleiss kappa of 0.51, lending our measure of creativity acceptable inter-rater reliability. The final measure of whether an idea was good or not was computed by thresholding the number of votes, so that good = 1 if at least $k$ judges rated it as good. We report results for both liberal and strict settings $k=2,3$.


\xhdr{Evaluation}
For $k=2$, out of $749$ total ideas collected, $249$ ideas were judged as good by this measure. As mentioned above, we use the Fisher-Yates shuffle to assign seeds to conditions. To take a conservative approach, as a first step we look only at seeds that appeared across all three conditions ($9$ such seeds), to put the conditions on par with one another. By this slicing of the data, there were $208$ good ideas. The proportion of good ideas in our condition was $46\%$ ($N=105$). Next  was the random baseline with $37\%$ ($49$), and finally the TF-IDF baseline achieved $30\%$ (N=$54$). These results are significant by a $\chi^2$ proportion test ($p \le .01$). We thus observe that both in terms of the absolute number of positively-rated ideas and in terms of proportions, our approach was able to generate a considerably large relative positive effect, leading to better ideas.

For $k=3$ (the majority vote), out of $749$ total ideas collected, $184$ ideas were judged as good. Again, we start by looking only at seeds that appeared across all three conditions ($9$ such seeds). This leaves $154$ good ideas. The proportion of good ideas in our condition was $38\%$ ($N=118$). Next-up was the random baseline with $22\%$ ($68$), and finally the TF-IDF baseline achieved $21\%$ (N=$63$), with $p<.01$. By looking at the more conservative majority-vote threshold, the observed effect of our method only increases.

Looking only at seeds that appeared across all conditions was a basic way to make sure we cancel out possible confounding factors. A more refined way is attempting to model these effects and condition on them, as follows.

We are interested in the likelihood that a given idea is good, or pr(good), as a function of inspiration condition. However, ideas are not independent: each participant generated multiple ideas, and ideas were proposed for different seeds. Failing to account for these dependencies would lead to inaccurate estimates of the effects of the inspirations: some participants may be better at generating ideas than others, while some seeds might be more easy/difficult than others. Therefore, we used a generalized linear mixed model, with a fixed effect of inspiration condition, and random effects of participant and seed (to model within-participant and within-seed dependencies between ideas). 

For $k=2$, our resulting model (with fixed effect of inspiration condition) yields a significant reduction in variance compared to a null model with no fixed effects, Likelihood ratio $\chi^2(3) = 67.96$, $p < .01$. The model also yields a reduction in Akaike Information Criterion (AIC), from $682.28$ in the null model to $620.32$, indicating that the improved fit to the data is not due to overfitting.

For $k=3$, the model also yields a significant drop in variance compared to a null model, Likelihood ratio $\chi^2(3) = 92.38$, $p < .01$, with AIC  dropping from $682.28$ in the null model to $595.90$.

\begin{figure}[t!]
\includegraphics[width=0.75\linewidth]{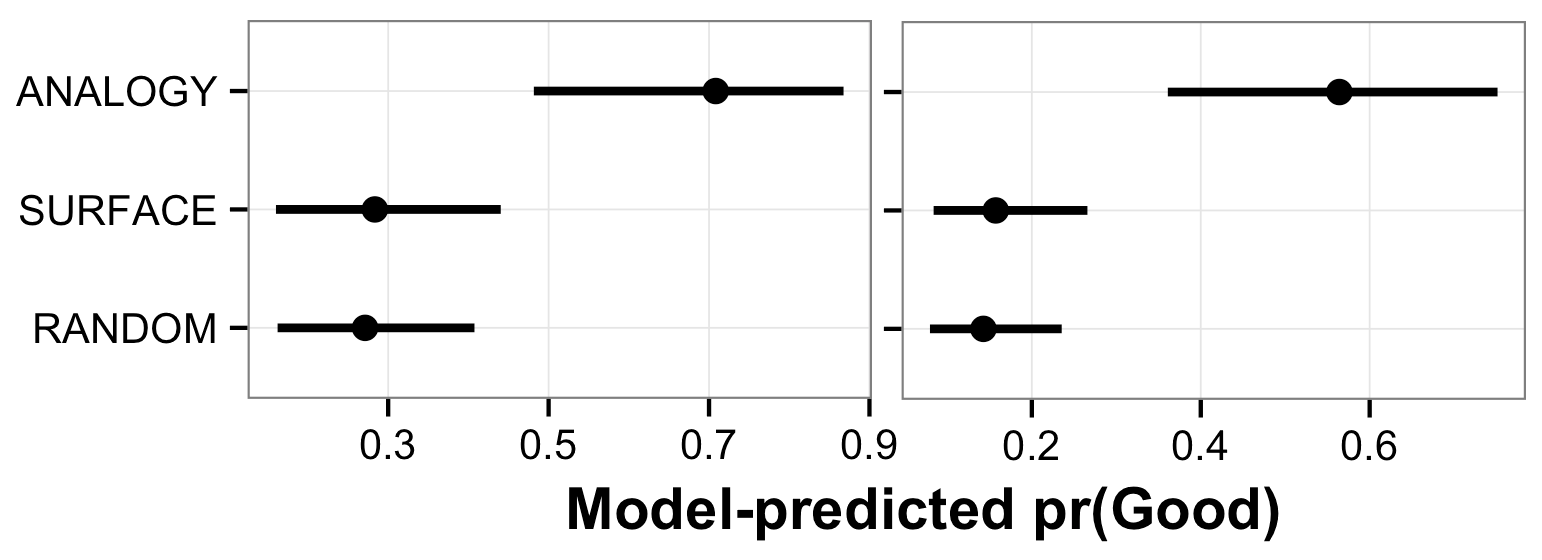}
        \caption{\small  Showing proportion estimates by our random-effect logistic regression, for $k=2$ (left) and $k=3$ (right). Participants are significantly more likely to generate good ideas for the redesign ideation task when given inspirations from our analogy approach compared to baseline-surface and baseline-random approaches\label{fig:creativity}}
\end{figure}

As Figure \ref{fig:creativity} shows, our method led to a significantly higher probability for good ideas. For $k=2$, pr(Good) = $0.71$, $95\%$ confidence interval = [$0.48$, $0.87$] in our condition. TF-IDF had pr(Good) = $0.28$ [$0.16$, $0.44$], and random had pr(Good) = $0.27$ [$0.16$, $0.41$]. The advantages of the analogy condition over each baseline are both substantial and statistically significant, B = $-1.81$, $p < .01$ vs. TF-IDF, and B = $-1.88$, $p < .01$ vs. random. For $k=3$, we had pr(Good) = $0.56$, [$0.36$, $0.75$]. TF-IDF had pr(Good) = $0.16$ [$0.08$, $0.27$], and random had pr(Good) = $0.14$ [$0.08$, $0.24$], B = $-1.94$, $p < .01$ vs. TF-IDF, and B = $-2.05$, $p < .01$ vs. random.

 Note that confidence intervals for the probability estimates are relatively wide (more so, unsurprisingly, for $k=2$). Replications of this experiment, possibly with more data, could yield results somewhere in between, with more precise estimates on the true size of the effect. The main take-away of this study is that our approach yields a reliable increase in participants' creative ability.
 
 \remove{suggest that the difference between the analogy and baseline conditions could be as large as an approximately $4$x increase (considering the lower tip of the confidence intervals for the baselines, and the upper tip of the confidence interval for the analogy condition), or as small as approximately a 10\% increase (considering the upper tip of the confidence intervals for the baselines, and the lower tip of the confidence interval for the analogy condition).}

%% file: Discussion.tex
In this paper, we sought to develop a scalable approach to finding analogies in large, messy, real-world datasets. We explored the potential of learning and leveraging a weak structural representation (i.e., purpose and mechanism vectors) for product descriptions. We leverage crowdsourcing techniques to construct a training dataset with purpose/mechanism annotations, and use an RNN to learn purpose and mechanism vectors for each product. We demonstrate that these learned vectors allow us to find analogies with higher precision than traditional information-retrieval similarity metrics like TF-IDF, LSA, GloVe and LDA.

Our ideation evaluation experiment further illustrates the effectiveness of our approach: participants had a higher likelihood of generating good ideas for the redesign ideation task when they received inspirations sampled by our analogy approach (tuned to be similar in purpose, but different in mechanism), compared to a traditional (TF-IDF) baseline or random sampling approach. From a psychological perspective, the benefits of our inspirations are likely due to our approach's superior ability to sample diverse yet still structurally similar inspirations, since diversity of examples is a known robust booster for creative ability \cite{chan_importance_2015}. The TF-IDF approach yielded inspirations likely to be relevant, but also likely to be redundant and homogeneous, while the random sampling approach yields diversity but not relevance.


While moving to a ``weak'' structural representation based on purpose and mechanism significantly increased the feasibility of analogy-finding, extensions may be necessary to generalize to other domains besides product descriptions. For example, our purpose and mechanism vectors did not distinguish between higher and lower level purposes/mechanisms, or core/peripheral purposes/ mechanisms, and also did not encode dependencies between particular purposes/mechanisms. These are potentially fruitful areas for future work and may be especially important when moving from relatively simple product descriptions to more complex data such as scientific papers, in which purposes and mechanisms can exist at multiple hierarchical levels (e.g., ``accelerate innovation'' vs. ``learn a vector representation of the purpose of an item''). More generally, we believe exploring the tradeoffs between degree of structure, learnability (including costs of generating training data, accuracy, and generalizability) and utility for augmenting innovation could lead to interesting points in the design space that could have both theoretical and practical value.